\documentclass[10pt, a4paper]{article}
\usepackage{lrec-coling2024} 
\usepackage{amsmath}

\usepackage{natbib}
\usepackage{multibib}
\makeatletter
\def\@mb@citenamelist{cite,citep,citet,citealp,citealt,citepalias,citetalias}
\makeatother
\newcites{languageresource}{~}

\usepackage{graphicx}
\usepackage{tabularx}
\usepackage{soul}
\usepackage{titlesec}
\titleformat{\section}{\normalfont\large\bfseries\center}{\thesection.}{1em}{}
\titleformat{\subsection}{\normalfont\SmallTitleFont\bfseries\raggedright}{\thesubsection.}{1em}{}
\titleformat{\subsubsection}{\normalfont\normalsize\bfseries\raggedright}{\thesubsubsection.}{1em}{}
\renewcommand\thesection{\arabic{section}}
\renewcommand\thesubsection{\thesection.\arabic{subsection}}
\renewcommand\thesubsubsection{\thesubsection.\arabic{subsubsection}}

\usepackage{xcolor}
\usepackage{hyperref}
 \definecolor{darkblue}{rgb}{0, 0, 0.5}
  \hypersetup{colorlinks=true, citecolor=darkblue, linkcolor=darkblue, urlcolor=darkblue}

\usepackage{xstring}

\usepackage{color}

\usepackage[framemethod=tikz]{mdframed}

\title{Conversational Grounding: Annotation and Analysis of Grounding Acts and Grounding Units}

\name{Biswesh Mohapatra\textsuperscript{1}, Seemab Hassan\textsuperscript{2}, Laurent Romary\textsuperscript{1}, Justine Cassell\textsuperscript{1,3}} 

\address{\textsuperscript{1}Inria, \textsuperscript{2}University of Stuttgart, \textsuperscript{3}Carnegie Mellon University \\
         \{Biswesh.Mohapatra, Laurent.Romary, Justine.Cassell\}@inria.fr, seemabhassan97@gmail.com\\ }

\abstract{
Successful conversations often rest on common understanding, where all parties are on the same page about the information being shared. This process, known as conversational grounding, is crucial for building trustworthy dialog systems that can accurately keep track of and recall the shared information.  The proficiencies of an agent in grounding the conveyed information significantly contribute to building a reliable dialog system. Despite recent advancements in dialog systems, there exists a noticeable deficit in their grounding capabilities. Traum \cite{Traumphd} provided a framework for conversational grounding introducing Grounding Acts and Grounding Units, but substantial progress, especially in the realm of Large Language Models, remains lacking. To bridge this gap, we present the annotation of two dialog corpora employing Grounding Acts, Grounding Units, and a measure of their degree of grounding. We discuss our key findings during the annotation and also provide a baseline model to test the performance of current Language Models in categorizing the grounding acts of the dialogs. Our work aims to provide a useful resource for further research in making conversations with machines better understood and more reliable in natural day-to-day collaborative dialogs.
 \\ \newline  }

\begin{document}

\maketitleabstract

\section{Introduction}
In linguistics, Clark and Brennan \cite{Clark1991-CLAGIC} proposed the concept of "common ground" which is a shared body of knowledge and assumptions collectively forged by the participants in a conversation. This common ground, accumulated progressively, isn't just crafted through verbal exchanges, but also through various other modalities. For instance, gestures, like pointing at surrounding objects, signaling acknowledgment with a nod, or eye contact to solicit more information, play vital roles, as pointed out by Nakano et al. 
 \cite{nakano-etal-2003-towards}. This interactive process of building common ground during a conversation, by making sure that both interlocutors share an understanding of the information that is being exchanged, is called 'Conversational Grounding'. It is within this realm of engagement that participants navigate, negotiate, and dissipate the inherent uncertainties in dialogs, thereby ensuring its effectiveness. Uncertainty may be resolved by providing additional context, like "the big one next to the Ferrari," or by requests for clarification, such as "You mean the blue one?". it is therefore essential to include a grounding mechanism in dialog systems, whether the  system assumes the role of  speaker — sensing a lack of comprehension from the listener and supplementing with additional information, or as the listener - requesting clarifications when deemed necessary to move forward with the conversation.  \cite{skantze-dogruoz-2023-open, benotti, skantz2022} further discuss the need for common ground in today's dialog systems.

The introduction of Common Ground in conversation \cite{Clark1991-CLAGIC, Clark1989ContributingTD} laid a foundation for subsequent explorations into the multifaceted aspects of grounding, both in human-human and human-machine dialog. For the latter, numerous efforts have addressed grounding challenges in conversational systems, with a notable focus on rule-based modular dialog systems. The complexity arises from the inherently dynamic nature of dialogs, which cannot merely be treated as strings of grammatically correct texts. For example, overlapping utterances may refer to different pieces of information, with their grounding therefore occurring out of  order. In addition, interlocutors don't always manifest overt signs of having grounded, and as a result, interlocutors may need to discern signs of understanding from their counterparts or ask if information has been understood.

In order to address these challenges, Traum \cite{Traumphd} proposed the concept of Grounding Acts (GA) and Common Grounding units (CGU), that provide a way to discretize the phenomenon of grounding into basic units. A CGU represents the unit of conversation in which grounding takes placeand is the smallest information unit of common ground. A CGU is composed of several individual utterances. Each individual utterance has a corresponding GA. These GAs help in the creation of the CGUs. Figures \ref{fig:ga_example} and \ref{fig 1: annotation} provide an example of these grounding acts. We discuss these GAs and CGUs in more detail later in Section \ref{GA/CGU}. 

\begin{figure}
    \begin{mdframed}[
        linecolor=black,linewidth=0.5pt,%
        frametitlerule=true,%
        apptotikzsetting={\tikzset{mdfframetitlebackground/.append style={%
            shade,left color=white, right color=yellow!20}}}, 
        frametitlerulecolor=black,
        frametitlerulewidth=1pt, innertopmargin=\topskip,
        frametitle={Grounding Acts of a CGU},
        outerlinewidth=1.25pt
    ]
    \textbf{User1} : And I see one stone only  - \textcolor{blue}{Initiate} \\
    \textbf{User2} : Only one? - \textcolor{blue}{Request-Repair} \\
    \textbf{User1} : yes one big stone - 
     \textcolor{blue}{Repair} \\
    \textbf{User2} : Okay yeah - \textcolor{blue}{Explicit-Acknowledgment} \\
    \end{mdframed}

\caption{Example of Grounding Acts from the 'Spot the Difference' dataset of a particular CGU starting from initiate and grounding with an acknowledgment}
\label{fig:ga_example}
\end{figure}

Once the dialogs have been grouped into CGUs using the GAs, it becomes easier to extract the correct information from these CGUs, store them, and use them effectively during the course of the conversation. The pioneering work by Traum and Allen \cite{TraumGA} delved into GAs and used them to form CGUs in the the Trains-93 Dataset \cite{Trains}. Subsequent investigations into CGUs predominantly adhered to the annotation guidelines laid out by Nakatani and Traum \cite{nakatani}. While they assert CGUs as fundamental units, they don't provide  information for the annotation of the GAs. Much of the work around CGUs has been analytical and the ones that have been implemented as working computational models, like \cite{Trains},  employ GAs as a fundamental unit to develop CGUs. 

In this paper, we concurrently annotate the GAs and CGUs to help us mitigate ambiguity, facilitating a more coherent annotation process for our annotators. We elucidate the challenges encountered during the annotation of such natural day-to-day collaborative dialogs containing the kind of back-and-forth conversations required for minimizing ambiguity. This annotation leads us to propose adding new grounding categories inspired from \cite{Roque2008DegreesOG}. We then evaluate the ability of a T5 model to categorize utterances into Grounding Acts, further grouping them into Common Grounding Units, thereby establishing a baseline for forthcoming investigations using large language models. 

While works like \cite{bunt-etal-2020-iso} have provided an annotation scheme, we chose to base our annotation on the work of \cite{TraumGA} since GAs and CGUs provides a more granular structure to the dialogs, thereby aiding in advancing research on conversational grounding, especially in the context of task-oriented dialog where the breadth of conversational topics/domains is not restricted. Such task-based dialogs which require back-and-forth information exchange on unanticipated domains or referents are very common in day-to-day life and need to be studied with respect to Conversational Grounding in order to build better dialog systems.

We annotate two existing openly available dialog corpora with GAs, CGUs and the degree of grounding at each stage. These corpora are - 1) Meetup Dataset \cite{meetup} 2) Spot the Difference \cite{spotthedifference}. While meetup is a written chat corpus, Spot the Difference is spoken dialogue. Both, however,  We work with both in order to highlight the distinctions between types of dialog. Both, however, contain conversational grounding negotiations.

We make available these large corpora annotated with GAs, CGUs and degree of grounding, to support future research. Finally, we provide a coding manual for GAs and CGUs which includes the changes made to \cite{Traumphd}'s classification in order to facilitate annotation and take into account the specificities of natural task-oriented dialog of this kind. 

\section{Related Work}
The importance of robust conversational grounding capabilities in dialogue systems has led to a lively literature. In linguistics, \cite{Clark1991-CLAGIC} examined the inherent uncertainty in dialogs that interlocutors negotiate and eliminate during the grounding process, identifying four distinct states of uncertainty:
\begin{enumerate}
    \item B didn't notice that A uttered any utterance u. 
    \item B noticed that A uttered some u.
    \item B correctly heard u.
    \item B understood what A meant by u.
\end{enumerate}

Paek and Horvitz \cite{PaekandHorvitz}'s subsequent work  proposed different hierarchical levels of mutual understanding:  Channel, Signal, Intention and Conversation. The collective insights from these works \cite{PaekandHorvitz, Clark1991-CLAGIC} show the importance of a hierarchy of understanding to truly solve the issue of conversational grounding in the spoken dialog system. Traum \cite{Traumphd} provided another hierarchical structure of Grounding Acts and Common Grounding Units as discussed in section \ref{GA/CGU}.
Later theories like Centering Theory \cite{grosz-etal-1983-providing} and Domain Reference theory \cite{DRT} offered methods to represent and store the grounded information. However, their effectiveness was limited to closed domains, largely due to their heavy reliance on rule-based approaches. Therefore, efforts towards developing more adaptable models that can categorize utterances into various grounding units irrespective of the domain would significantly contribute to the field.

Numerous datasets have been curated to aid research endeavors in conversational grounding. Among these, Photobook \cite{photobook} is a large-scale collection of visually-grounded, task-oriented dialogs in English designed to permit the study of shared dialog history accumulated during conversation. Two participants are asked to identify shared images in their respective photo books by exchanging text messages. OneCommon and Dynamic OneCommon \cite{Udagawa_Aizawa_2019, Udagawa2021MaintainingCG} provided data necessitating enhanced common grounding capabilities amidst continuous and partially observable contexts. Talk The Walk \cite{talk_the_walk} amalgamated three crucial aspects: perception, enabling a tourist to observe the world; action, facilitating navigation through the environment for the tourist; and interactive dialog, aiding both the tourist and a guide in accomplishing their shared objective. They constructed a virtual 2D grid environment by manually capturing 360-views of several neighborhoods of NYC. Although the aforementioned datasets predominantly encapsulate text-based dialogs, HCRC Maptask \cite{hcrc} emerged as one of the pioneering datasets comprising spoken dialogs, designed explicitly for conversational grounding studies. Here, two individuals are presented with maps, where one possesses a route, and the other is tasked with reproducing that route post-discussion. 
A  limitation across many of these datasets is that a substantial portion of the utterances comprise merely one or two words, as participants strive to fulfill the specified task quickly instead of conveying and discussing their thoughts. This scenario hardly mirrors the more organic conversations typically occurring between interlocutors in day-to-day conversations. In contrast, datasets like Teach \cite{teach}, present a more specific scenario conducive to developing agents receptive to commands. Here, a designated "teacher" imparts instructions, such as coffee preparation, to a "follower," and throughout this process, references to various objects are made. This setup, however, remains tailored to a particular scenario and, hence may not capture the essence of conversational grounding in more generalized dialog. Challenges such as the Give Challenge \cite{gargett-etal-2010-give} tried to remove some of the constraints present in other datasets. However, such challenges are hard to conduct due to logistical limitations. We describe our rationale for selecting our specific datasets in section \ref{datasets}.

Several studies \cite{Mushin2003DiscourseSG, traum96_icslp, takeoka} have explored CGUs, but with the aim of analyzing CGUs alongside other phenomena such as intonation, dialog acts (which differ from grounding acts), boundary tones, etc., and therefore, were not conducted on a broad scale. Furthermore, many of these studies utilized the annotation coding manual from \cite{nakatani}, which accounted for CGUs as fundamental units but didn't look at GAs. Additionally, the efforts aimed at developing computational models of CGUs \cite{Traumphd, Traum1999ComputationalMO, Traum-incremental} employed GAs as the foundational unit for constructing CGUs and encompassed merely 1000 utterances from the Trains-93 dataset \cite{Trains}. The datasets we deploy here include 4934 utterances in Spot the Difference and 5131 utterances in Meetup dataset.

\section{Grounding Acts and Common Grounding Units}
\label{GA/CGU}

Traum \cite{TraumGA}'s introduction of Grounding Acts facilitated the decomposition of the phenomenon into fundamental units. It includes the following categories -

\begin{enumerate}
\item \textbf{Initiate}: An initial utterance of a grounding unit which proposes information to be grounded;

\item \textbf{Continue}: Continuation of the previous act from the same speaker. Separate intonation phrase but syntactically and semantically part of the same act;

\item \textbf{Acknowledge}: An acknowledgment of the proposed information from the interlocutor;

\item \textbf{Repair}: Correction of previously uttered material or addition of omitted material that will change the listener's interpretation of the speaker's communicative intention. It is different from domain clarifications where the interlocutor brings back a domain/topic and re-opens it. Repairs  concern the grounding of content and not changes to previously grounded content;

\item \textbf{Request Repair}: Often distinguished from repair and acknowledgment using intonation where  where the interlocutor asks for further clarification;

\item \textbf{Request Acknowledge}: Attempt to elicit an acknowledgement of the previous utterance;

\item \textbf{Cancel}: Closes off the current Discourse Unit ungrounded. 

\end{enumerate}

 The GAs and CGUs described here provided a way to create the hierarchical level of understanding discovered in previous work. The seven categories of GAs help us understand when a piece of information is considered grounded. Each unit or CGU starts with an initiate and is grounded after an implicit or explicit form of acknowledgment. Later, Roque and Traum \cite{Roque2008DegreesOG} emphasized that the Grounding Acts could also help us to understand the degree of uncertainty in a dialog.

\section{Datasets}
\label{datasets}

Upon reviewing the available datasets, we have decided to utilize the Meetup \cite{meetup} and Spot the Difference datasets \cite{spotthedifference} for our annotations. 

The Meetup dataset featured a scenario wherein two participants are placed on a 2D grid, with each vertex representing a room. The objective is for the participants to converge on the same room, despite only having visibility of their respective rooms. Navigational actions (east, west, north, or south) move participants to new rooms, unveiling the image of the newly entered room. Achieving the goal necessitates room descriptions, formulation and communication of a converging strategy, retention of room descriptions shared by the counterpart, and mental modeling of the other participant's room configurations. Although the dataset is text-based, it serves as a great resource for exploring and developing grounding models. Unlike many tasks that designate a leader role to one participant, this task creates an egalitarian dynamic where both participants can assume initiator or responder roles. Consequently, we plan to continue to use this dataset for our experiments.

The "Spot the Difference" dataset presents a spoken dialog scenario involving two participants, each provided with slightly distinct images. Positioned in separate rooms, they communicate via audio calls to identify the discrepancies between their respective images. This dataset iss a valuable asset for examining conversational grounding in a spoken dialog framework. 

"Spot the difference" and Meetup datasetare therefore ideal for evaluating and implementing grounding models. We annotated all 430 dialogs from the Meetup dataset, containing 5131 utterances. However, due to the vast size of the Spot the Difference dataset, we only annotated 50 dialogs chosen at random, comprising 4934 utterances. The link for annotated data is in the footnotes \footnote{\href{https://osf.io/qfcnm/?view_only=34e7259fe8fc4ade82d55ba7d5105ffe}{Link for data and annotation manual}}.

\begin{figure*}[ht]
\caption{Annotation of meetup dataset}
\label{fig 1: annotation}
\centering
\includegraphics[width=0.7\textwidth]{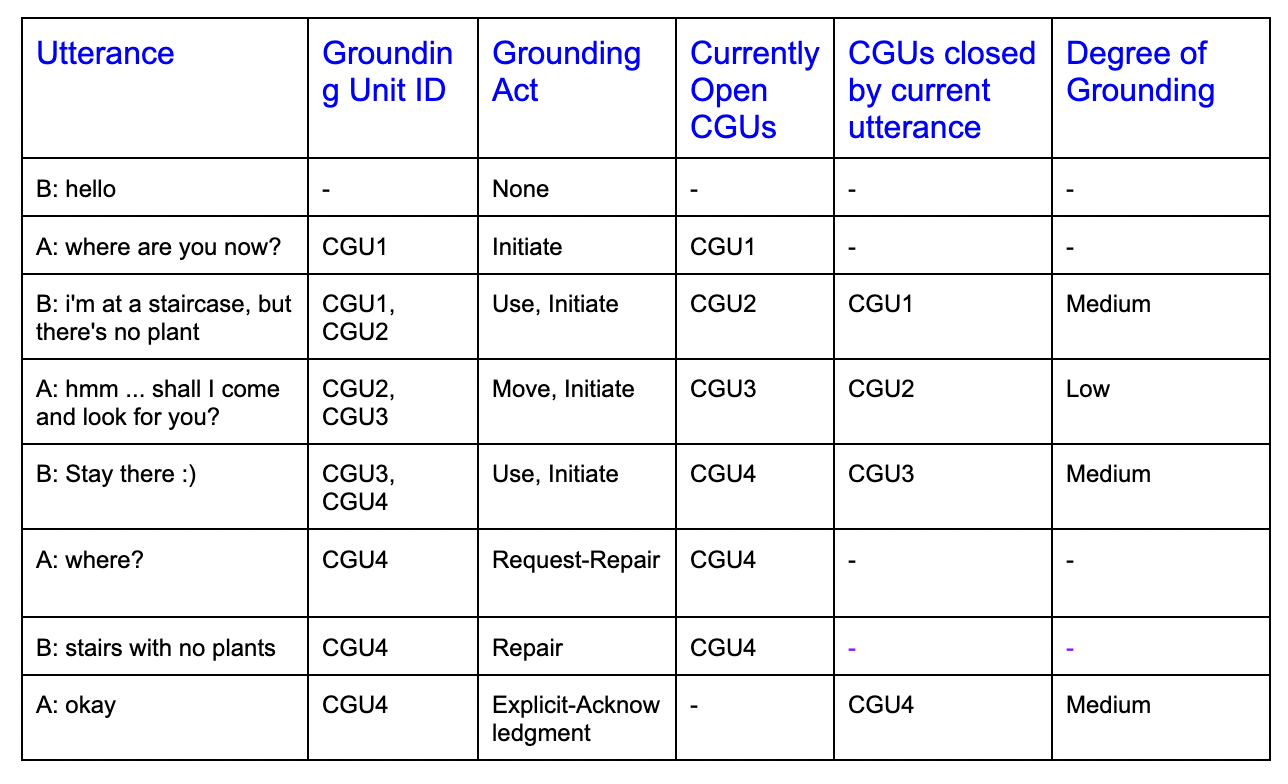}
\end{figure*}

\section{Annotation Scheme}

Annotation was carried out by two individuals possessing a background in computational linguistics. An Inter-Rater Reliability (IRR) score of 0.81 and 0.78 was attained on the annotation of the Meetup dataset and Spot the Difference dataset respectively, using Cohen's Kappa, indicating a substantial level of agreement using a well-detailed annotation manual. As the CGUs could be calculated easily once the GAs were calculated, we have provided IRRs for GAs here. The annotation process, proved to offer valuable insights into the workings of conversational grounding in such natural, day-to-day dialogs, as discussed in Section \ref{challenges} and Section \ref{analysis}.

In the annotation task, besides the categories of acts outlined by \cite{TraumGA}, we propose the inclusion of two additional categories: Repeat and None. The Repeat category encompasses instances where the speaker reiterates previously mentioned information, while the None category was necessary to account for many utterances that did not significantly contribute to the grounding process. Furthermore, drawing inspiration from Roque and Traum \cite{Roque2008DegreesOG}, we introduce sub-categories of acknowledgment to offer a more nuanced understanding of that dynamic - 

\begin{enumerate}
    \item \textbf{Explicit Acknowledgment}: describes utterances that indicate acknowledgment regarding the provided information, such as 'okay', 'ya', etc.
    \item \textbf{Repeat Back}: used when the listener repeats the information provided.
    \item \textbf{Move on}: necessary when the listener transitions to a different topic, and thus implicitly signifies acknowledgment, but only in cases where the preceding information did not necessitate a response. Questions or directive statements requiring action are examples where transitioning away without addressing them would not imply acknowledgment.
    \item \textbf{Use}: characterizes the listener utilizing the shared information to provide additional details. For example, if the speaker says "I will be going to Paris next month" and the listener replies with "Oh I love Paris!!", then it's an implicit form of grounding of the speaker's information from the listener. 
\end{enumerate}

The inclusion of these sub-categories aided the annotators in delineating the boundaries between the CGUs more effectively. The 'Use' category further facilitated the establishment of connections between different CGUs related to the same topic. This created a hierarchical structure that could further be used for both reasoning and storage of the grounded information in dialog systems.

During the annotation process, we regarded information as grounded when it was acknowledged by the listener, whether explicitly or implicitly. However, note that a Common Grounding Unit (CGU) may be reopened after being grounded in instances of a Repair, a Request-Repair, or a Request-Acknowledge. Moreover, a CGU can be removed from the list of grounded CGUs if the proposer revokes it.

In our annotations, we also assessed the level of grounding for each Common Grounding Unit (CGU) upon its grounding, categorizing them into four levels: High, Medium, Low, and Ambiguous. CGUs that are re-affirmed by the speaker through a repeat-back are assigned a high degree of grounding. Meanwhile, those grounded by the 'Use' grounding act or 'Explicit Acknowledgment' are designated a medium degree. When a CGU is grounded due to a 'Move on' act, it's assigned a low degree, indicating a higher likelihood of alteration in future discussions. Lastly, certain utterances appeared to be grounded, yet annotators found them to be ambiguous, placing them in the 'ambiguous' category. For instance, in a scenario from the Meetup dataset, a speaker poses multiple questions about a room, such as 'Does it have a chair? Is there a doll on the table as well? Do they have refrigerators?' and the user provides an ambiguous answer by saying 'No it's not my room'. In such cases, it remains unclear which, if any, of the three utterances the ambiguous answer is addressing. While further clarifications were required, the participants moved to other information in the original dialog. Hence, we decided to keep such utterances under the 'ambiguous' category even if the CGU was closed. It is noteworthy that in some instances, especially within the spoken dialog corpora, the degree of grounding was determined based on the context, including prosody, and the intent conveyed in the utterance.

Figure \ref{fig 1: annotation} shows an example of the annotations. The GAs help decide the CGU IDs of the utterance. After each utterance, we maintain a list of all the CGUs that haven't been grounded yet under the \textit{Open CGUs} column and also mention the CGUs that were closed by that particular utterance in the \textit{CGUs Closed} column.  Each CGU ID is closed/grounded once it is implicitly or explicitly acknowledged. Following their grounding, we also designate the respective degree. As illustrated in the figure, it's possible for a single utterance to be associated with multiple CGUs.

\section{Challenges}
\label{challenges}

\begin{figure}
    \begin{mdframed}[
        linecolor=black,linewidth=0.5pt,%
        frametitlerule=true,%
        apptotikzsetting={\tikzset{mdfframetitlebackground/.append style={%
            shade,left color=white, right color=green!20}}}, 
        frametitlerulecolor=black,
        frametitlerulewidth=1pt, innertopmargin=\topskip,
        frametitle={Continue vs Repair Example},
        outerlinewidth=1.25pt,
        nobreak=true,
    ]
    
            \textcolor{blue}{User A: And it has like two shapes} - \textbf{Initiate} \\
            \textcolor{red}{User B: Yeah} - \textbf{Exp-Acknowledgment} \\
            \textcolor{blue}{User A: like two polygon shapes} - \textbf{Continue/Repair}\\ 
            \textcolor{red}{User B: yeah} - \textbf{Exp-Acknowledgment} \\
    \end{mdframed}
\vspace{-2mm}
\caption{Instance of confusion between Continue and Repair in Spot the Difference dataset}
\label{fig:continue/repair}
\end{figure}

\begin{figure}
    \begin{mdframed}[
        linecolor=black,linewidth=0.5pt,%
        frametitlerule=true,%
        apptotikzsetting={\tikzset{mdfframetitlebackground/.append style={%
            shade,left color=white, right color=green!20}}}, 
        frametitlerulecolor=black,
        frametitlerulewidth=1pt, innertopmargin=\topskip,
        frametitle={Cancelation of Request-Repair Example},
        outerlinewidth=1.25pt,
        nobreak=true,
    ]
    
        \textcolor{blue}{A: Is there a bed?} - \textbf{Initiate} (CGU 1) \\
	\textcolor{red}{B: yes} - \textbf{Use} (CGU 1) \\
	\textcolor{blue}{A: A chair?} - \textbf{Initiate} (CGU 2) \\
	\textcolor{red}{B: yes} - \textbf{Use} (CGU 2) \\
	\textcolor{blue}{A: Was there a bed again?} - \textbf{Request-Repair} (CGU 1) \\
        \textcolor{blue}{A: ah yes, never mind} - \textbf{Cancel} (CGU 1)

    \end{mdframed}
\vspace{-2mm}
\caption{Instance of Cancelation of a re-opened CGU leading to grounding}
\label{fig:reqrepair-cancel}
\end{figure}

During the course of annotation, we encountered various scenarios that posed challenges to the process. For instance, there were instances where distinguishing between a 'Continue' and a 'Repair' was challenging. The provided definitions occasionally fell short, necessitating a deeper contextual understanding for accurate categorization of the utterances. Figure \ref{fig:continue/repair} illustrates an instance where User A, having been interrupted by User B, appears to repair the information by adding newer information. However, upon closer listening, it becomes clear that it was a continuation of the original statement.

Additionally, we faced complications with back-channel responses, as they weren't always indicative of acknowledgments, but sometimes simply indicated that the interlocutor was listening, and hence, were annotated as 'None'. They were examples of the early levels of grounding in the four-level hierarchy provided by \cite{Clark1991-CLAGIC}. Similarly, within the spoken dialog corpus, we discovered instances where the listener was either continuing or repairing the utterance originated by the initiator, adding another layer of complexity to annotation. 

For spoken dialog, determining whether an utterance was merely the interlocutor murmuring, verbalizing thoughts, or engaging in a conversation with the other participant presented a challenge. These instances were particularly hard to annotate as they occasionally lacked coherence when evaluated solely based on the transcript, without auditory context. This underscores the significance of the cues contained in audio, a topic that we are currently pursuing. 

In the written dialog, we observed instances where a repair was initiated on a closed CGU but later the repair itself was canceled. However, such cancellations do not result in the dismissal of the entire CGU (Common Grounding Unit). Hence, a cancellation does not always remove a CGU but can also ground it, as can be seen in Figure \ref{fig:reqrepair-cancel}. In \textbf{23} instances within the Meetup dataset, the annotators found it necessary to revise annotation of a specific utterance upon examination of subsequent utterances. We denote such instances with a '*' in the annotated corpus. Additionally, in the written dialogs, there were places where both participants were typing simultaneously, potentially overlooking the utterances from their counterparts. To address this issue, we computed the average response time for utterances of different lengths (in number of words).   We subsequently employed the average response time to ascertain whether it was feasible for the interlocutor to have read, comprehended, and grounded the preceding utterance, and then write the subsequent utterance. If the response time was less then we did not consider the preceding utterance to be grounded.

We also came across examples containing "non-linear" or "disordered" utterances which made it challenging to group the utterances into CGUs. Figure \ref{fig:disordered} shows an example where the question was asked by the participant before providing the entire context. Similarly, Figure \ref{fig:multi-thread} shows a multi-threaded dialog where User 1 responds with a "No you are not" to User 2's "me too" utterance.

\begin{figure}
    \begin{mdframed}[
        linecolor=black,linewidth=0.5pt,%
        frametitlerule=true,%
        apptotikzsetting={\tikzset{mdfframetitlebackground/.append style={%
            shade,left color=white, right color=red!20}}}, 
        frametitlerulecolor=black,
        frametitlerulewidth=1pt, innertopmargin=\topskip,
        frametitle={Disordered dialog Example},
        outerlinewidth=1.25pt,
        nobreak=true,
    ]
    
         \textcolor{blue}{Original Dialog} \\
            \textbf{User A}: Blue curtains \\
            \textbf{User A}: Am I in it? \\
            \textbf{User A}: Pink floor \\

        \textcolor{blue}{More fluent order} \\
            \textbf{User A}: Blue curtains \\
            \textbf{User A}: Pink floor \\
            \textbf{User A}: Am I in it? 
    \end{mdframed}
\vspace{-2mm}
\caption{Example of disordered dialog in Meetup}
\label{fig:disordered}
\end{figure}

\begin{figure}
    \begin{mdframed}[
        linecolor=black,linewidth=0.5pt,%
        frametitlerule=true,%
        apptotikzsetting={\tikzset{mdfframetitlebackground/.append style={%
            shade,left color=white, right color=red!20}}}, 
        frametitlerulecolor=black,
        frametitlerulewidth=1pt, innertopmargin=\topskip,
        frametitle={Multi-threaded Example},
        outerlinewidth=1.25pt,
        nobreak=true,
    ]
    
          \textcolor{blue}{\textbf{User 1} : I am in the reading room.} \\
          \textcolor{teal}{\textbf{User 2} : me too.} \\
          \textcolor{teal}{\textbf{User 2} : 2 people on computers in the background.} \\
          \textcolor{blue}{\textbf{User 1} : Book shelf on the wall.} \\
          \textcolor{blue}{\textbf{User 1} : One table and a chair.} \\
          \textcolor{blue}{\textbf{User 1} : No you are not.} \\ 
    \end{mdframed}
\vspace{-4mm}
\caption{Example of Multi-threaded dialog}
\label{fig:multi-thread}
\end{figure}
\vspace{-2mm}



\section{Analysis}
\label{analysis}

The aggregate counts of Grounding Acts across various categories are presented in Table \ref{table:meetup_ga}, facilitating an examination of frequently occurring acts versus the seldom occurring ones. This distinction is valuable as current Language Learning Models (LLMs), which largely depend on their training data, might struggle to grasp certain phenomena associated with the less frequent acts. Likewise, Table \ref{table:std_ga} enumerates the counts for the 'Spot the Difference' dataset. It is evident that 'Spot the Difference' possesses a notably higher count of explicit Acknowledgments relative to 'Move', highlighting the variation in acknowledgment methods in across contexts. Moreover, while there is a significant presence of Repairs and Repeat-Backs in 'Spot the Difference', acts such as Cancel are minimally represented in both scenarios.

\begin{table}[h]
\centering
\caption{Total number of instances of each Grounding Act in Meetup Dataset \\}
\label{table:meetup_ga}
\begin{tabular}{|l|c|c|}
\hline
\textbf{Grounding Act}           & \textbf{\# Instances} & \textbf{\% Instances}\\  \hline
Initiate                  & 2633        & 49.03 \\ \hline
Cancel                  & 4        & 0.07 \\ \hline
Explicit Ack.           & 364      & 6.77  \\ \hline
Move                    & 937      & 17.44 \\ \hline
Repair                  & 86       & 1.60 \\ \hline
Repeat                  & 21       & 0.39 \\ \hline
Repeat-Back             & 10       & 0.18 \\ \hline
Request-Ack.            & 1        & 0.01 \\ \hline
Request-Repair          & 42       & 0.78 \\ \hline
Use                     & 1273     & 23.70 \\ \hline
\end{tabular}
\end{table}


\begin{table}[h]
\centering
\caption{Total number of instances of each Grounding Act in Spot the Difference Dataset \\}
\label{table:std_ga}
\begin{tabular}{|l|c|c|}
\hline
\textbf{Grounding Act}           & \textbf{\# Instances} & \textbf{\% Instances} \\  \hline
Initiate                  & 3723        & 62.28 \\ \hline
Cancel                  & 2        & 0.03 \\ \hline
Explicit Ack. & 1233     & 20.62 \\ \hline
Move                    & 117      & 1.95 \\ \hline
Repair                  & 977      & 16.3 \\ \hline
Repeat                  & 124      & 2.07 \\ \hline
Repeat-Back             & 153      & 2.55 \\ \hline
Request-Ack.            & 3        & 0.05 \\ \hline
Request-Repair          & 339      & 5.67 \\ \hline
Use                     & 542      & 9.06 \\ \hline
\end{tabular}
\end{table}

In the process of annotating CGUs for the Meetup dataset, we identified \textbf{27} instances where a specific CGU was revisited post-grounding. Among these, \textbf{11} instances exhibited a time-lapse exceeding 10 seconds, with the longest gap stretching to \textbf{1 minute 9 seconds} — notable given that conversations in the dataset never exceeded 3 minutes. For the Spot the Difference dataset, in 630 instances the CGUs were revisited post-grounding with the longest gap stretching to \textbf{1 minute 38 seconds}.

While a majority of CGUs in Meetup were grounded in the following utterance (1599 instances to be precise), the longest grounding trajectory spanned \textbf{13} utterances from initiation until acknowledgment. In "Spot the Difference", the longest trajectory was even more pronounced, encompassing \textbf{85} utterances. This observation underscores the necessity for dialog systems to handle long-term contexts and the capability to revise them.  

Furthermore, we noted \textbf{32} instances in Meetup where CGUs were tagged as ambiguous, highlighting the need for dialog models to be able to discuss and dismiss ambiguities when necessary. "Spot the Difference", on the other hand, did not contain such ambiguities, perhaps because of the potential for prosody to disambiguate.

Additionally, in "Spot the Difference,"  \textbf{5} instances were found where participants were murmuring to themselves, rather than to their interlocutor. These annotations are identified to aid future analysis and modeling efforts. Likewise, in \textbf{171} instances in Spot the Difference, subsequent utterances provided context essential for accurate annotation. These instances are flagged with a '*'. This emphasizes the need for models to be capable of considering subsequent utterances as context.  Similarly, in some places the interlocutors spoke at the same time, preventing accurate transcription. These have been flagged with a '\#'.

\section{Model}
\begin{figure}
    \begin{mdframed}[
        linecolor=black,linewidth=0.5pt,%
        frametitlerule=true,%
        apptotikzsetting={\tikzset{mdfframetitlebackground/.append style={%
            shade,left color=white, right color=green!20}}}, 
        frametitlerulecolor=black,
        frametitlerulewidth=1pt, innertopmargin=\topskip,
        frametitle={T5 input Example},
        outerlinewidth=1.25pt,
        nobreak=true,
    ]
    
         Context: \\
         \text{[00:15]} User1: I see a lamp [CGU 1] \\
         \text{[00:17]} User1: go west [CGU 2]\\
         
         First utterance of context(U1) belongs to CGU 1 and Second utterance(U2) belongs to CGU 2 \\
         
         Next Utterance(U3) : "[00:19] A: no lamp here" \\

         Final Input: \\ 
         
         For CGU1 - \\ 
         \textcolor{red}{<special\_token>}\textcolor{blue}{U1}\textcolor{red}{<special\_token>}</s>\textcolor{blue}{U2}\\
         </s></s>\textcolor{purple}{U3}</s>\textcolor{olive}{Use}</s> \\

         For CGU2 - \\ 
         \textcolor{blue}{U1}</s>\textcolor{red}{<special\_token>}\textcolor{blue}{U2}\textcolor{red}{<special\_token>}
         </s></s>\textcolor{purple}{U3}</s>\textcolor{olive}{None}</s> \\

         For CGU3 - \\
        \textcolor{blue}{U1}</s>\textcolor{blue}{U2}</s></s>\textcolor{purple}{U3}</s>\textcolor{olive}{Initiate}</s>        
    \end{mdframed}
\vspace{-2mm}
\caption{Example of input to T5 model}
\label{fig:input_model}
\end{figure}

\begin{table}[h]
\centering
\caption{T5 performance on Meetup dataset}
\vspace{5pt}
\label{Meetup-result}
\begin{tabular}{|l|c|c|}
\hline
GA           & \begin{tabular}[c]{@{}l@{}}Accuracy w/o \\ weighted loss\end{tabular} & \begin{tabular}[c]{@{}l@{}}Accuracy with \\ weighted loss\end{tabular} \\ \hline
Use:         & 0.66                                                                  & 0.74                                                                   \\ \hline
Move:        & 0.68                                                                  & 0.70                                                                   \\ \hline
Req-Ack      & 0.00                                                                  & 0.00                                                                   \\ \hline
Req-Repair   & 0.00                                                                  & 0.00                                                                   \\ \hline
Repair       & 0.00                                                                  & 0.57                                                                   \\ \hline
Initiate     & 0.99                                                                  & 0.99                                                                   \\ \hline
Repeat       & 0.00                                                                  & 0.00                                                                   \\ \hline
Explicit-Ack & 0.62                                                                  & 0.11                                                                   \\ \hline
Continue     & 0.00                                                                  & 0.05                                                                   \\ \hline
Repeat-Back  & 0.00                                                                  & 0.00                                                                   \\ \hline
Cancel       & 0.00                                                                  & 0.00                                                                   \\ \hline
\end{tabular}
\end{table}

\begin{table}[h]
\centering
\caption{T5 performance on Spot the Difference dataset \\}
\label{STD-result}
\begin{tabular}{|l|c|c|}
\hline
GA           & \begin{tabular}[c]{@{}l@{}}Accuracy w/o \\ weighted loss\end{tabular} & \begin{tabular}[c]{@{}l@{}}Accuracy with \\ weighted loss\end{tabular} \\ \hline
Use:         &  0.00                                                                     & 0.01                                                                   \\ \hline
Move:        &  0.00                                                                     & 0.00                                                                   \\ \hline
Req-Ack      &  0.00                                                                     & 0.00                                                                   \\ \hline
Req-Repair   &   0.00                                                                    & 0.00                                                                   \\ \hline
Repair       &  0.00                                                                     & 0.00                                                                   \\ \hline
Initiate     &   0.23                                                                    & 0.31                                                                   \\ \hline
Repeat       &   0.00                                                                    & 0.00                                                                   \\ \hline
Explicit-Ack &  0.35                                                                     & 0.86                                                                   \\ \hline
Continue     &    0.15                                                                   & 0.24                                                                   \\ \hline
Repeat-Back  &   0.00                                                                    & 0.00                                                                   \\ \hline
Cancel       &  0.00                                                                     & 0.00                                                                   \\ \hline
\end{tabular}
\end{table}

Following the annotation of Grounding Acts (GAs) and Conversational Grounding Units (CGUs), we proceeded to evaluate the capability of current transformer-based models \cite{transformer} to categorize these utterances into various Grounding Act categories. Given that CGUs are derived based on the GAs, there is no need for two distinct models. CGUs were created with 'Initiate' and grounded at an 'Acknowledgment'. We decided to use the T5 \cite{T5}, a text-to-text encoder-decoder-based transformer model. We divided the utterances into sets using the ratio of 70:15:15 for train, dev, and test sets, respectively, each containing same proportion of every GA. Since an utterance can belong to multiple CGUs, we needed to calculate the possible classification of that utterance for every CGU. If an utterance did not belong to a CGU then we classified it as None for that CGU. The model's input representation is depicted in Figure \ref{fig:input_model}. For CGU 1, the subsequent utterance is designated a GA of 'Use', whereas, for CGU 2, it is assigned 'None'. Concurrently, this utterance also initiates a new CGU 3. The input to the model contains the dialog history, the next utterance, and the label each separated by a separator token '</s>'. For CGU 1, the utterance U1 of the context is surrounded with a special token to mark that the utterance belongs to the focal CGU. Similarly, for CGU 2, the special token is around utterance U2. For CGU 3, since no utterance in dialog history belongs to this new CGU, we do not mark any other with the special token. 

As described above, experiments used a T5-base model, fine-tuned with cross-entropy loss. The model's accuracy across various Grounding Acts (GAs) on the meetup dataset is detailed in Table \ref{Meetup-result}. We also outline the accuracy following the application of a weighted loss function during fine-tuning. While the adoption of weighted loss enhanced the model's accuracy, the model performance is not substantial. Thus, we cobsuder ut ti serve as a baseline reference for subsequent research efforts on modeling GAs. Similarly, we show an improved model performance on the Spot the Difference dataset in Table \ref{STD-result}. A probable explanation for the observed performance discrepancy across distinct categories could be the significant imbalance in category sizes, as illustrated in Tables \ref{table:meetup_ga} and \ref{table:std_ga}. While the small number of examples in these datasets could be seen as problematic, we justify their use by pointing out that they are representative of natural conversations of this sort. 

\section{Conclusion}

In this paper, we investigated the domain of conversational grounding,  highlighting the pivotal role of Grounding Acts (GAs) and Common Grounding Units (CGUs) in structuring dialog. The use of Meetup and Spot the Difference datasets underscored diverse grounding challenges while also illuminating the potential of rigorously annotated GAs and CGUs as instrumental tools for advancing research in this domain. The structured categorization paved the way for more effective extraction, storage, and utilization of dialogic information, essential for building an effective conversational agent. We also employed a T5 model to categorize utterances into GAs and subsequently group them into CGUs, establishing a preliminary benchmark for future computational endeavors in this sphere. In the future, it will be interesting to compare the performance of models trained on GAs and those trained directly on CGUs. In conclusion, we believe that the large annotated corpora we make available, along with a refined coding manual, provide a robust foundation for subsequent investigations, extending a valuable resource for conversational grounding research. 

\nocite{*}
\section{Bibliographical References}\label{reference}

\bibliographystyle{lrec-coling2024-natbib}
\bibliography{lrec-coling2024-example}

\begin{thebibliography}{43}
\expandafter\ifx\csname natexlab\endcsname\relax\def\natexlab#1{#1}\fi

\bibitem[{Allen et~al.(1994)Allen, Schubert, Ferguson, Heeman, Hwang, Kato, Light, Martin, Miller, Poesio, and Traum}]{Trains}
James~F. Allen, Lenhart~K. Schubert, George Ferguson, Peter~A. Heeman, Chung~Hee Hwang, Tsuneaki Kato, Marc Light, Nathaniel~G. Martin, Bradford~W. Miller, Massimo Poesio, and David~R. Traum. 1994.
\newblock \href {https://api.semanticscholar.org/CorpusID:18225253} {The trains project: a case study in building a conversational planning agent}.
\newblock \emph{J. Exp. Theor. Artif. Intell.}, 7:7--48.

\bibitem[{Axelsson et~al.(2022)Axelsson, Buschmeier, and Skantze}]{skantz2022}
Agnes Axelsson, Hendrik Buschmeier, and Gabriel Skantze. 2022.
\newblock \href {https://doi.org/10.3389/fcomp.2022.744574} {Modeling feedback in interaction with conversational agents—a review}.
\newblock \emph{Frontiers in Computer Science}, 4.

\bibitem[{Bara et~al.(2021)Bara, CH-Wang, and Chai}]{mindcraft}
Cristian-Paul Bara, Sky CH-Wang, and Joyce Chai. 2021.
\newblock \href {https://doi.org/10.18653/v1/2021.emnlp-main.85} {{M}ind{C}raft: Theory of mind modeling for situated dialogue in collaborative tasks}.
\newblock In \emph{Proceedings of the 2021 Conference on Empirical Methods in Natural Language Processing}, pages 1112--1125, Online and Punta Cana, Dominican Republic. Association for Computational Linguistics.

\bibitem[{Benotti and Blackburn(2021)}]{benotti}
Luciana Benotti and Patrick Blackburn. 2021.
\newblock \href {https://doi.org/10.18653/v1/2021.eacl-main.41} {Grounding as a collaborative process}.
\newblock In \emph{Proceedings of the 16th Conference of the European Chapter of the Association for Computational Linguistics: Main Volume}, pages 515--531, Online. Association for Computational Linguistics.

\bibitem[{Bunt et~al.(2020)Bunt, Petukhova, Gilmartin, Pelachaud, Fang, Keizer, and Pr{\'e}vot}]{bunt-etal-2020-iso}
Harry Bunt, Volha Petukhova, Emer Gilmartin, Catherine Pelachaud, Alex Fang, Simon Keizer, and Laurent Pr{\'e}vot. 2020.
\newblock \href {https://aclanthology.org/2020.lrec-1.69} {The {ISO} standard for dialogue act annotation, second edition}.
\newblock In \emph{Proceedings of the Twelfth Language Resources and Evaluation Conference}, pages 549--558, Marseille, France. European Language Resources Association.

\bibitem[{Clark and Brennan(1991)}]{Clark1991-CLAGIC}
Herbert~H. Clark and Susan~E. Brennan. 1991.
\newblock Grounding in communication.
\newblock In Lauren Resnick, Levine B., M.~John, Stephanie Teasley, and D., editors, \emph{Perspectives on Socially Shared Cognition}, pages 13--1991. American Psychological Association.

\bibitem[{Clark and Schaefer(1989)}]{Clark1989ContributingTD}
Herbert~H. Clark and Ed~Schaefer. 1989.
\newblock Contributing to discourse.
\newblock \emph{Cogn. Sci.}, 13:259--294.

\bibitem[{de~Vries et~al.(2018)de~Vries, Shuster, Batra, Parikh, Weston, and Kiela}]{talk_the_walk}
Harm de~Vries, Kurt Shuster, Dhruv Batra, Devi Parikh, Jason Weston, and Douwe Kiela. 2018.
\newblock Talk the walk: Navigating new york city through grounded dialogue.
\newblock \emph{ArXiv}, abs/1807.03367.

\bibitem[{Denis(2010)}]{DRT}
Alexandre Denis. 2010.
\newblock \href {https://aclanthology.org/W10-4203} {Generating referring expressions with reference domain theory}.
\newblock In \emph{Proceedings of the 6th International Natural Language Generation Conference}. Association for Computational Linguistics.

\bibitem[{DeVault and Stone(2009)}]{DeVault2009LearningTI}
David DeVault and Matthew Stone. 2009.
\newblock Learning to interpret utterances using dialogue history.
\newblock In \emph{Conference of the European Chapter of the Association for Computational Linguistics}.

\bibitem[{Dimosthenis and Joakim(2021)}]{Kontogiorgos2021}
Kontogiorgos Dimosthenis and Gustafson Joakim. 2021.
\newblock \href {https://doi.org/10.3389/fpsyg.2021.623657} {Measuring collaboration load with pupillary responses - implications for the design of instructions in task-oriented hri}.
\newblock Front Psychol.

\bibitem[{Frank and Goodman(2012)}]{RSA2012}
Michael~C. Frank and Noah~D. Goodman. 2012.
\newblock \href {https://doi.org/10.1126/science.1218633} {Predicting pragmatic reasoning in language games}.
\newblock \emph{Science}, 336(6084):998--998.

\bibitem[{Fried et~al.(2021)Fried, Chiu, and Klein}]{fried-etal-2021-reference}
Daniel Fried, Justin Chiu, and Dan Klein. 2021.
\newblock \href {https://doi.org/10.18653/v1/2021.emnlp-main.163} {Reference-centric models for grounded collaborative dialogue}.
\newblock In \emph{Proceedings of the 2021 Conference on Empirical Methods in Natural Language Processing}, pages 2130--2147, Online and Punta Cana, Dominican Republic. Association for Computational Linguistics.

\bibitem[{Gargett et~al.(2010)Gargett, Garoufi, Koller, and Striegnitz}]{gargett-etal-2010-give}
Andrew Gargett, Konstantina Garoufi, Alexander Koller, and Kristina Striegnitz. 2010.
\newblock \href {http://www.lrec-conf.org/proceedings/lrec2010/pdf/532_Paper.pdf} {The {GIVE}-2 corpus of giving instructions in virtual environments}.
\newblock In \emph{Proceedings of the Seventh International Conference on Language Resources and Evaluation ({LREC}'10)}, Valletta, Malta. European Language Resources Association (ELRA).

\bibitem[{Grosz et~al.(1983)Grosz, Joshi, and Weinstein}]{grosz-etal-1983-providing}
Barbara~J. Grosz, Aravind~K. Joshi, and Scott Weinstein. 1983.
\newblock \href {https://doi.org/10.3115/981311.981320} {Providing a unified account of definite noun phrases in discourse}.
\newblock In \emph{21st Annual Meeting of the Association for Computational Linguistics}, pages 44--50, Cambridge, Massachusetts, USA. Association for Computational Linguistics.

\bibitem[{Haber et~al.(2019)Haber, Baumg{\"a}rtner, Takmaz, Gelderloos, Bruni, and Fern{\'a}ndez}]{photobook}
Janosch Haber, Tim Baumg{\"a}rtner, Ece Takmaz, Lieke Gelderloos, Elia Bruni, and Raquel Fern{\'a}ndez. 2019.
\newblock \href {https://doi.org/10.18653/v1/P19-1184} {The {P}hoto{B}ook dataset: Building common ground through visually-grounded dialogue}.
\newblock In \emph{Proceedings of the 57th Annual Meeting of the Association for Computational Linguistics}, pages 1895--1910, Florence, Italy. Association for Computational Linguistics.

\bibitem[{Ilinykh et~al.(2019)Ilinykh, Zarrie{\ss}, and Schlangen}]{meetup}
Nikolai Ilinykh, Sina Zarrie{\ss}, and David Schlangen. 2019.
\newblock \href {http://semdial.org/anthology/Z19-Ilinykh_semdial_0006.pdf} {Meet up! a corpus of joint activity dialogues in a visual environment}.
\newblock In \emph{Proceedings of the 23rd Workshop on the Semantics and Pragmatics of Dialogue - Full Papers}, London, United Kingdom. SEMDIAL.

\bibitem[{Kontogiorgos et~al.(2019)Kontogiorgos, Pereira, and Gustafson}]{Kontogiorgos2019}
Dimosthenis Kontogiorgos, Andre Pereira, and Joakim Gustafson. 2019.
\newblock \href {https://doi.org/10.1145/3340555.3353722} {Estimating uncertainty in task-oriented dialogue}.
\newblock In \emph{2019 International Conference on Multimodal Interaction}, ICMI '19, page 414–418, New York, NY, USA. Association for Computing Machinery.

\bibitem[{LAION-AI(2023)}]{openassistant}
LAION-AI. 2023.
\newblock Open-assistant.
\newblock \url{https://github.com/LAION-AI/Open-Assistant}.

\bibitem[{Lopes et~al.(2018)Lopes, Hemmingsson, and {\AA}strand}]{spotthedifference}
Jos{\'e} Lopes, Nils Hemmingsson, and Oliver {\AA}strand. 2018.
\newblock \href {https://aclanthology.org/L18-1305} {The spot the difference corpus: a multi-modal corpus of spontaneous task oriented spoken interactions}.
\newblock In \emph{Proceedings of the Eleventh International Conference on Language Resources and Evaluation ({LREC} 2018)}, Miyazaki, Japan. European Language Resources Association (ELRA).

\bibitem[{Mushin et~al.(2003)Mushin, Stirling, Fletcher, and Wales}]{Mushin2003DiscourseSG}
Ilana Mushin, Lesley Stirling, Janet Fletcher, and Roger Wales. 2003.
\newblock \href {https://api.semanticscholar.org/CorpusID:6076282} {Discourse structure, grounding, and prosody in task-oriented dialogue}.
\newblock \emph{Discourse Processes}, 35:1 -- 31.

\bibitem[{Nakano et~al.(2003)Nakano, Reinstein, Stocky, and Cassell}]{nakano-etal-2003-towards}
Yukiko Nakano, Gabe Reinstein, Tom Stocky, and Justine Cassell. 2003.
\newblock \href {https://doi.org/10.3115/1075096.1075166} {Towards a model of face-to-face grounding}.
\newblock In \emph{Proceedings of the 41st Annual Meeting of the Association for Computational Linguistics}, pages 553--561, Sapporo, Japan. Association for Computational Linguistics.

\bibitem[{Nakatani and Traum(2001)}]{nakatani}
Christine Nakatani and David Traum. 2001.
\newblock Coding discourse structure in dialogue (version 1.0).

\bibitem[{Ouyang et~al.(2022)Ouyang, Wu, Jiang, Almeida, Wainwright, Mishkin, Zhang, Agarwal, Slama, Ray, Schulman, Hilton, Kelton, Miller, Simens, Askell, Welinder, Christiano, Leike, and Lowe}]{instructGPT}
Long Ouyang, Jeff Wu, Xu~Jiang, Diogo Almeida, Carroll~L. Wainwright, Pamela Mishkin, Chong Zhang, Sandhini Agarwal, Katarina Slama, Alex Ray, John Schulman, Jacob Hilton, Fraser Kelton, Luke Miller, Maddie Simens, Amanda Askell, Peter Welinder, Paul Christiano, Jan Leike, and Ryan Lowe. 2022.
\newblock \href {http://arxiv.org/abs/2203.02155} {Training language models to follow instructions with human feedback}.

\bibitem[{Padmakumar et~al.(2022)Padmakumar, Thomason, Shrivastava, Lange, Narayan-Chen, Gella, Piramuthu, Tur, and Hakkani-Tur}]{teach}
Aishwarya Padmakumar, Jesse Thomason, Ayush Shrivastava, Patrick Lange, Anjali Narayan-Chen, Spandana Gella, Robinson Piramuthu, Gokhan Tur, and Dilek Hakkani-Tur. 2022.
\newblock {TEACh: Task-driven Embodied Agents that Chat}.
\newblock In \emph{Proceedings of the AAAI Conference on Artificial Intelligence}, volume~36, pages 2017--2025.

\bibitem[{Paek and Horvitz(2000)}]{PaekandHorvitz}
Tim Paek and Eric Horvitz. 2000.
\newblock Conversation as action under uncertainty.
\newblock In \emph{Proceedings of the Sixteenth Conference on Uncertainty in Artificial Intelligence}, UAI'00, page 455–464, San Francisco, CA, USA. Morgan Kaufmann Publishers Inc.

\bibitem[{Park et~al.(2023)Park, O'Brien, Cai, Morris, Liang, and Bernstein}]{park2023generative}
Joon~Sung Park, Joseph~C. O'Brien, Carrie~J. Cai, Meredith~Ringel Morris, Percy Liang, and Michael~S. Bernstein. 2023.
\newblock \href {http://arxiv.org/abs/2304.03442} {Generative agents: Interactive simulacra of human behavior}.

\bibitem[{Peng et~al.(2022)Peng, Galley, He, Brockett, Liden, Nouri, Yu, Dolan, and Gao}]{godel}
Baolin Peng, Michel Galley, Pengcheng He, Chris Brockett, Lars Liden, Elnaz Nouri, Zhou Yu, Bill Dolan, and Jianfeng Gao. 2022.
\newblock \href {https://www.microsoft.com/en-us/research/publication/godel-large-scale-pre-training-for-goal-directed-dialog/} {Godel: Large-scale pre-training for goal-directed dialog}.
\newblock arXiv.

\bibitem[{Radford et~al.(2021)Radford, Kim, Hallacy, Ramesh, Goh, Agarwal, Sastry, Askell, Mishkin, Clark et~al.}]{clips}
Alec Radford, Jong~Wook Kim, Chris Hallacy, Aditya Ramesh, Gabriel Goh, Sandhini Agarwal, Girish Sastry, Amanda Askell, Pamela Mishkin, Jack Clark, et~al. 2021.
\newblock Learning transferable visual models from natural language supervision.
\newblock In \emph{International conference on machine learning}, pages 8748--8763. PMLR.

\bibitem[{Raffel et~al.(2020)Raffel, Shazeer, Roberts, Lee, Narang, Matena, Zhou, Li, and Liu}]{T5}
Colin Raffel, Noam Shazeer, Adam Roberts, Katherine Lee, Sharan Narang, Michael Matena, Yanqi Zhou, Wei Li, and Peter~J. Liu. 2020.
\newblock Exploring the limits of transfer learning with a unified text-to-text transformer.
\newblock \emph{J. Mach. Learn. Res.}, 21(1).

\bibitem[{Roque and Traum(2008)}]{Roque2008DegreesOG}
Antonio Roque and David Traum. 2008.
\newblock \href {https://aclanthology.org/W08-0107} {Degrees of grounding based on evidence of understanding}.
\newblock In \emph{Proceedings of the 9th {SIG}dial Workshop on Discourse and Dialogue}, pages 54--63, Columbus, Ohio. Association for Computational Linguistics.

\bibitem[{Skantze and Do{\u{g}}ru{\"o}z(2023)}]{skantze-dogruoz-2023-open}
Gabriel Skantze and A.~Seza Do{\u{g}}ru{\"o}z. 2023.
\newblock \href {https://doi.org/10.18653/v1/2023.sigdial-1.57} {The open-domain paradox for chatbots: Common ground as the basis for human-like dialogue}.
\newblock In \emph{Proceedings of the 24th Annual Meeting of the Special Interest Group on Discourse and Dialogue}, pages 605--614, Prague, Czechia. Association for Computational Linguistics.

\bibitem[{Takeoka and Shimojima(2002)}]{takeoka}
Atsue Takeoka and Atsushi Shimojima. 2002.
\newblock \href {https://doi.org/10.3115/1118121.1118145} {Grounding styles of aged dyads: an exploratory study}.
\newblock In \emph{Proceedings of the 3rd SIGdial Workshop on Discourse and Dialogue - Volume 2}, SIGDIAL '02, page 188–195, USA. Association for Computational Linguistics.

\bibitem[{Thompson et~al.(1993)Thompson, Anderson, Bard, Doherty-Sneddon, Newlands, and Sotillo}]{hcrc}
Henry~S. Thompson, Anne Anderson, Ellen~Gurman Bard, Gwyneth Doherty-Sneddon, Alison Newlands, and Cathy Sotillo. 1993.
\newblock \href {https://aclanthology.org/H93-1005} {The {HCRC} map task corpus: Natural dialogue for speech recognition}.
\newblock In \emph{{H}uman {L}anguage {T}echnology: Proceedings of a Workshop Held at Plainsboro, New Jersey, March 21-24, 1993}.

\bibitem[{Touvron et~al.(2023)Touvron, Lavril, Izacard, Martinet, Lachaux, Lacroix, Rozi{\`e}re, Goyal, Hambro, Azhar, Rodriguez, Joulin, Grave, and Lample}]{llama}
Hugo Touvron, Thibaut Lavril, Gautier Izacard, Xavier Martinet, Marie-Anne Lachaux, Timoth{\'e}e Lacroix, Baptiste Rozi{\`e}re, Naman Goyal, Eric Hambro, Faisal Azhar, Aur'elien Rodriguez, Armand Joulin, Edouard Grave, and Guillaume Lample. 2023.
\newblock Llama: Open and efficient foundation language models.
\newblock \emph{ArXiv}, abs/2302.13971.

\bibitem[{Traum and Allen(1992)}]{TraumGA}
David Traum and James Allen. 1992.
\newblock \href {https://doi.org/10.21437/ICSLP.1992-41} {A "speech acts" approach to grounding in conversation}.
\newblock \emph{Proc. 2nd International Conference on Spoken Language Processing (ICSLP 1992)}.

\bibitem[{Traum(1999)}]{Traum1999ComputationalMO}
David~R. Traum. 1999.
\newblock \href {https://api.semanticscholar.org/CorpusID:2918832} {Computational models of grounding in collaborative systems}.

\bibitem[{Traum and Heeman(1996)}]{traum96_icslp}
David~R. Traum and Peter~A. Heeman. 1996.
\newblock \href {https://doi.org/10.21437/ICSLP.1996-484} {{Utterance units and grounding in spoken dialogue}}.
\newblock In \emph{Proc. 4th International Conference on Spoken Language Processing (ICSLP 1996)}, pages 1884--1887.

\bibitem[{Traum(1995)}]{Traumphd}
David~Rood Traum. 1995.
\newblock \emph{A Computational Theory of Grounding in Natural Language Conversation}.
\newblock Ph.D. thesis, USA.
\newblock UMI Order No. GAX95-23171.

\bibitem[{Udagawa and Aizawa(2019)}]{Udagawa_Aizawa_2019}
Takuma Udagawa and Akiko Aizawa. 2019.
\newblock \href {https://doi.org/10.1609/aaai.v33i01.33017120} {A natural language corpus of common grounding under continuous and partially-observable context}.
\newblock \emph{Proceedings of the AAAI Conference on Artificial Intelligence}, 33(01):7120--7127.

\bibitem[{Udagawa and Aizawa(2021)}]{Udagawa2021MaintainingCG}
Takuma Udagawa and Akiko Aizawa. 2021.
\newblock Maintaining common ground in dynamic environments.
\newblock \emph{Transactions of the Association for Computational Linguistics}, 9:995--1011.

\bibitem[{Vaswani et~al.(2017)Vaswani, Shazeer, Parmar, Uszkoreit, Jones, Gomez, Kaiser, and Polosukhin}]{transformer}
Ashish Vaswani, Noam Shazeer, Niki Parmar, Jakob Uszkoreit, Llion Jones, Aidan~N. Gomez, Lukasz Kaiser, and Illia Polosukhin. 2017.
\newblock \href {https://arxiv.org/pdf/1706.03762.pdf} {Attention is all you need}.

\bibitem[{Visser et~al.(2014)Visser, Traum, DeVault, and Akker}]{Traum-incremental}
Thomas Visser, David Traum, David DeVault, and Rieks Akker. 2014.
\newblock \href {https://doi.org/10.1007/s12193-013-0147-7} {A model for incremental grounding in spoken dialogue systems}.
\newblock \emph{Journal on Multimodal User Interfaces}, 8.

\end{thebibliography}




\end{document}